\documentclass[letterpaper, 10 pt, journal, twoside]{IEEEtran}
\usepackage[utf8]{inputenc} % allow utf-8 input
\usepackage[T1]{fontenc}    % use 8-bit T1 fonts
\usepackage{url}            % simple URL typesetting
\usepackage{booktabs}       % professional-quality tables
\usepackage{amsfonts}       % blackboard math symbols
\usepackage{nicefrac}       % compact symbols for 1/2, etc.
\usepackage{microtype}      % microtypography
\usepackage{xcolor}         % colors

\usepackage{anyfontsize}
\usepackage{amsmath}
\usepackage{amssymb}
\usepackage{wrapfig}
\usepackage{subfig}
\usepackage{caption}
\usepackage{color}

\usepackage{physics}
\usepackage{bm}

\usepackage{algorithm}
\usepackage{algorithmic}
\usepackage{graphicx}

\usepackage{ragged2e}
\usepackage{fp}

\usepackage{ifthen}

\usepackage{pgfplots}
\usepackage{tikz}
\usepackage{tikz-3dplot}

\usepackage{ragged2e}
\usepackage{fp}

\usepackage{ifthen}

\usepackage{tikz}
\usetikzlibrary{
shapes.geometric, 
arrows,
 calc,
 automata,
 positioning,
 decorations,
 decorations.text,
 patterns}
 
 \usetikzlibrary{pgfplots.statistics, pgfplots.colorbrewer} 
\usepackage{pgfplotstable}

\usetikzlibrary{patterns.meta}

\usetikzlibrary{mindmap,snakes}

\usepackage{multirow}

\usepackage{algorithm}
\usepackage{algorithmic}
\usepackage{graphicx}
\usepackage[english]{babel}
\usepackage{array}
\usepackage[export]{adjustbox}

\usepackage{nicefrac}

\usepackage{xcolor}

\usepackage{colortbl}
\usepackage{makecell}
\usepackage{pifont}

\usepackage{mathptmx}

\usepackage[colorlinks=true,
    linkcolor=red,
    linkbordercolor=red,
    urlcolor=mylinkcolor,
    urlbordercolor=white
    ]{hyperref}

\pgfplotsset{
    /pgfplots/ybar legend/.style={
    /pgfplots/legend image code/.code={%
       \draw[##1,/tikz/.cd,yshift=-0.25em]
        (0cm,0cm) rectangle (3pt,0.8em);},
   },
}

\newcommand{\cmark}{\ding{51}}%
\newcommand{\xmark}{\ding{55}}

\definecolor{mylinkcolor}{rgb}{0.005, 0.3, 0.7}
\definecolor{suppcolor}{rgb}{0.6, 0.0, 0.9}

\usepackage[font=footnotesize]{caption}

\colorlet{baseclr}{gray}
\colorlet{rwclr}{white!90!baseclr}

\colorlet{dotaclr}{black}
\colorlet{dotbclr}{red}

\newcommand{\detector}{FFD}

\title{High-Speed Detector For Low-Powered Devices In Aerial Grasping \vspace{-0.5ex}}

\author{Ashish~Kumar$^{\dagger}$, Laxmidhar~Behera$^{\dagger}$, \IEEEmembership{Senior Member, IEEE} \vspace{-2.75ex}  \\
\thanks{Manuscript received: November 23, 2023; Accepted February 27, 2024. This paper was recommended for publication by Editor Hyungpil Moon upon evaluation of the Associate Editor and Reviewers' comments.}
\thanks{$^{\dagger}$EE, Indian Institute of Technology (IIT), Kanpur, India.
{\tt\small \{ashishkumar822@gmail.com,lbehera@iitk.ac.in\}}\\
\textbf{Supplementary:} \textcolor{mylinkcolor} {\footnotesize \url{https://github.com/ashishkumar822/\detector}}}
\thanks{Digital Object Identifier (DOI): see top of this page.}
}

\begin{document}

\maketitle

% % %
%%%%%%%%%%%%%%%%%%%%%%%%%%%%%%%%%%%%%%%%%%%%%%%%%%%%%%%%%%%%%%%%%%%%%%%%%%%%%%%%%%%%%%%
\begin{justify}

\begin{abstract}
Autonomous aerial harvesting is a highly complex problem because it requires numerous interdisciplinary algorithms to be executed on mini low-powered computing devices. Object detection is one such algorithm that is compute-hungry. In this context, we make the following contributions: (\textit{i}) Fast Fruit Detector (\detector), a resource-efficient, single-stage, and postprocessing-free object detector based on our novel latent object representation (\texttt{LOR}) module, query assignment, and prediction strategy. \detector{} achieves $\mathbf{100}$FPS$@$FP$\mathbf{32}$ precision on the latest $\mathbf{10}$W NVIDIA Jetson-NX embedded device while co-existing with other time-critical sub-systems such as control, grasping, SLAM, a major achievement of this work. (\textit{ii}) a method to generate vast amounts of training data without exhaustive manual labelling of fruit images since they consist of a large number of instances, which increases the labelling cost and time. (\textit{iii}) an open-source fruit detection dataset having plenty of very small-sized instances that are difficult to detect. Our exhaustive evaluations on our and MinneApple dataset show that FFD, being only a single-scale detector, is more accurate than many representative detectors, e.g. FFD is better than single-scale Faster-RCNN by $\mathbf{10.7}$AP, multi-scale Faster-RCNN by $\mathbf{2.3}$AP, and better than latest single-scale YOLO-v$\mathbf{8}$ by $\mathbf{8}$AP and multi-scale YOLO-v$\mathbf{8}$ by $\mathbf{0.3}$ while being considerably faster. 
%%
%\section*{\textcolor{suppcolor}{Supplementary Material}}
%
%
%\noindent
%\textbf{Code:} \textcolor{mylinkcolor}{\url{https://github.com/ashishkumar822/ADPD}} \\
%\textbf{Multimedia:}~\href{https://drive.google.com/file/d/1aVsJJlTRKAHqGhMx_ovG6dwlxgcGF6ZW/view?usp=share_link}{\textcolor{mylinkcolor}{Full Length Video}} \\

%\noindent
%\textbf{Code:} \textcolor{gray}{Will be released post acceptance.} \\
%\textbf{Multimedia:}~\href{https://drive.google.com/file/d/1IR9Zumkbg0oJM25IASfb8zOMWfVU8lLI/view?usp=share_link}{\textcolor{mylinkcolor}{Full Length Video}} \\
%
% \par
% \vspace{1ex}
% \noindent
% \textbf{\textcolor{mylinkcolor}{Code \& Dataset:}} \textcolor{gray}{\footnotesize \text{\url{https://github.com/ashishkumar822/FFD}}}\\
% \textbf{\textcolor{mylinkcolor}{Video:}}~\textcolor{gray}{See attachment.}

\end{abstract}

\end{justify}

\vspace{-0.75ex}
\IEEEpeerreviewmaketitle

\vspace{-3.00ex}

\section{Introduction}
\label{sec:intro}
Harvesting process in agriculture is a manpower-intensive and industrially important task, demanding high precision. With the rising applications of UAVs in agriculture, we foresee a huge scope of UAV-based grasping in the harvesting process. If one can harness the flying and maneuvering capabilities of UAVs, harvesting can continue $24\times7$ while significantly reducing the production costs, in outdoor orchards or recently emerged indoor vertical farming or precision agriculture.
\par
However, developing a UAV-based fully autonomous harvesting system is not as straightforward as combining several algorithms and then deploying. It is because such a system should work in constrained and GPS-denied workspaces with entirely onboard computations, which in turn requires several algorithms/sub-systems to work in conjunction \cite{towards}. Such algorithms mainly include object detection, tracking, positioning system, control system, and grasping system, and running all of them at desired rates altogether on a low-powered, computationally limited device is a bottleneck. However, we believe that if each sub-system can be optimized as per the task requirements, the above issue can be resolved. 
\par
In harvesting automation, object detection is both crucial and a compute-intensive task. Although modern deep learning-based detectors offer high accuracy and parallelization, their high computational demands pose an issue for low-powered devices. It is so because other sub-systems also require a certain amount of computing to be run at desired rates. In addition, a high frame processing rate of the detector is also desired by the control system in order to perform visual servoing to accurately reach and grasp a target object \cite{towards}.
\input{figures/intro_figure}%
\par
Motivated by this, we translate the object detection problem into re-innovating the head of a detector, because it consists of most of the hand-tuned hyperparameters and time-consuming post-processing steps apart from the backbone. As a result, we propose \textit{Fast-Fruit-Detector} (\textit{FFD}) inspired by the hyperparameters and post-processing free design of recent Detection-Transformer (DETR) \cite{detr}, while incorporating the task-centric observations from fruit harvesting, i.e. detection of fruits which appear smaller in images ($< 15 \times 15$ pixels).
\par
FFD represents objects as queries which are obtained by our novel \textit{Latent Object Representation} (\texttt{LOR}) module, directly from the backbone output instead of learning them \cite{detr}. These queries are used by our novel query assignment and matching strategy during the training phase. This turns FFD quite fast, accurate, resource-efficient, and postprocessing free while being a CNN-only design, free of compute-hungry Transformers. To the best of our knowledge, such speed and accuracy in the context of low-powered inference and robotic applications are still not visible in the literature. Despite we target FFD for fruits, it can be used in similar robotics applications. Summarily, main contributions of the paper are:
 \begin{enumerate}
\item Single-stage and postprocessing free detector, achieving $100$FPS$@$FP$32$ on $10$W NVIDIA Jetson-NX (Sec.~\ref{sec:method}).
\item A data multiplication approach to generate vast amounts of labelled training data from a small dataset (Sec.~\ref{sec:occ}).
\item A challenging fruit detection dataset (Sec.~\ref{sec:dataset}).
 \end{enumerate}
\par
Next, we discuss related works, followed by FFD and the data multiplication approach. Experiments are described in Sec.~\ref{sec:exp}, and Sec.~\ref{sec:conc} provides conclusions on the paper.

\section{Related Work}
\label{sec:rel}
\subsection{Convolutional Neural Network Based Detection}
\label{sec:rel_cnnobj}
RCNN \cite{rcnn} fused traditional selective search for region proposal and CNN to obtain box and classification score. Fast-RCNN \cite{fastrcnn} proposed RoI-pooling to convert proposal features into a fixed size, thus improving both the speed and accuracy over RCNN. Then to avoid CPU-intensive and sluggish region proposal step, Faster-RCNN \cite{fasterrcnn} proposed Region Proposal Network (RPN) and anchor boxes. RPN produces proposals as objectness score and coarse boxes relative to a huge number of anchors ($\sim20000$).
\par
However, Faster-RCNN training becomes two-staged, complex, and has hand-crafted steps and hyperparameters to handle issues such as matching ground-truth boxes with a large number of anchors, class imbalance due to fewer positive anchors (object), and large negative anchors (non-object), positive-negative ratio for box-mining that consumes computing resources due to its CPU-only execution \cite{fastrcnn, ssd}. This causes the accuracy and the runtime to be sensitive to the hyperparameter choices, thus necessitating hyperparameter tuning for a particular dataset which is a tedious process.
\par
Further, the large number of anchor boxes produces high confidence for an object, resulting in redundant detections. NMS handles this issue via an intersection-over-union (IoU) threshold, however, it often discards small objects due to their low prediction confidence and as they occupy very small regions in the feature map. Therefore such objects are detected at high-resolution feature maps, but since these maps lack large context, multi-scale detection via feature fusion \cite{fpn} is performed \cite{fasterrcnn, ssd}. It improves the accuracy but at the cost of increased run-time due to the processing of many anchors.
\par
YOLO \cite{yolov8}, SSD \cite{ssd} speed-up the inference but at the cost of reduced accuracy by eliminating RPN, however, box-matching, postprocessing and multi-scale detection remain intact. FCOS \cite{fcos} proposes an anchorless solution, however, postprocessing and feature fusion still exist. Moreover, mere backbone modifications \cite{cslyolo} or using depthwise separable convolutions in them \cite{tinydsod} does not help, mainly because of the fundamental design limitations, i.e. anchor boxes, NMS, multi-stage detection which still remain in the picture.
\par
The above limitations are bottlenecks in our case, i.e. the post-processing runtime overhead, and detecting small objects via FPN since fruits appear as small objects in the images.
\subsection{Transformer Based Object Detector}
Recent Detection-Transformer (DETR) \cite{detr} translates object detection into a set prediction problem while avoiding postprocessing and hyperparameters entirely. DETR first encodes input image using a CNN, which is fed to a Transformer module, and then predicts a priori fixed number of objects via a Feed Forward Neural Network (FFN). DETR is simpler relative to the CNN-based detectors, however, its transformer blocks are a bottleneck for embedded computing devices both in terms of memory and computing resources. In addition, it suffers from slower convergence which limits its direct deployment in our case. Nevertheless, its design strongly motivates the development of the proposed detector FFD.
\subsection{Detection For Fruit Harvesting}
The application of object detection in agriculture automation is huge. \cite{roy2016surveying} uses traditional feature-based vision for yield estimation. \cite{sa2016deepfruits} uses Faster-RCNN for vegetable and fruit detection. \cite{bargoti2017deep} again uses Faster-RCNN for apple detection in orchards and mentions the importance of having a fast and accurate detector. \cite{roy2019vision} uses Gaussian-Mixture-Model (GMM) for counting and yield mapping in apple orchards.
\par
Notably, these works employ existing detectors directly but do not focus on the detector design and improvements. As this is a fundamental requirement in this sector, we develop FFD for limited computing scenarios.

\section{Fast Fruit Detector}
\label{sec:method}
\begin{figure}[t]
\centering
\subfloat{\label{fig:fasterrcnn}}
\subfloat{\label{fig:ssd}}
\subfloat{\label{fig:detr}}
\subfloat{\label{fig:ftfd}}
\begin{tikzpicture}

%\FPeval{\imh}{10.5}
%\FPeval{\imw}{10.5}
%
%\FPeval{\imx}{48}
%\FPeval{\imy}{0-10.7}

%
\node (boundary) [draw=white!50!gray,  fill=white!98!black,rounded corners=0.2ex, minimum width=54ex, minimum height=12.5ex, xshift=20ex,yshift=-0.0ex]{};

\node (fasterrcnn) [draw=none, fill=none, scale=0.45]
{
\tikz
{
\node (cnn) [fill=none, draw=cyan, rounded corners=0.2ex]{CNN};
\node (rpn) [fill=none, draw=red, rounded corners=0.2ex, xshift=-3ex, yshift=-5ex]{RPN};
\node (nms) [fill=none, draw=orange!70!black, rounded corners=0.2ex, xshift=-4ex, yshift=-11.5ex]{\shortstack{Anchor box, \\ NMS}};
\node (roi) [fill=none, draw=green, rounded corners=0.2ex, xshift=1ex, yshift=-18.0ex]{RoI Pool};
\node (box) [fill=none, draw=gray, rounded corners=0.2ex, xshift=-4ex, yshift=-23ex]{Box};
\node (class) [fill=none, draw=gray, rounded corners=0.2ex, xshift=6ex, yshift=-23ex]{Class};
\draw [->, very thick] (cnn) -| ($(rpn.north) - (1ex, 0ex)$);
\draw [->, very thick] (rpn) -| ($(nms.north) - (5ex, 0ex)$);
\draw [->, very thick] (cnn) -| ($(roi.north) + (3.5ex, 0ex)$);
\draw [->, very thick] ($(nms.south) + (2ex, 0ex)$) -- ($(roi.north) - (3ex, 0ex)$);
\draw [->, very thick] (roi) -| ($(box.north) - (1ex, 0ex)$);
\draw [->, very thick] (roi) -| ($(class.north) + (1ex, 0ex)$);
}
};

\node (ssd) [draw=none, fill=none, scale=0.45, xshift=34ex]
{
\tikz
{
\node (cnn) [fill=none, draw=cyan, rounded corners=0.2ex]{CNN};
\node (box) [fill=none, draw=gray, rounded corners=0.2ex, xshift=-4ex, yshift=-10ex]{Box};
\node (class) [fill=none, draw=gray, rounded corners=0.2ex, xshift=4ex, yshift=-10ex]{Class};
\node (nms) [fill=none, draw=orange!70!black, rounded corners=0.2ex, xshift=-2ex, yshift=-21ex]{\shortstack{Anchor box, \\ NMS}};
\draw [->, very thick] (cnn) -| ($(box.north) - (1ex, 0ex)$);
\draw [->, very thick] (cnn) -| ($(class.north) + (1ex, 0ex)$);
\draw [->, very thick] (box) -- ($(nms.north) - (2ex, 0ex)$);
}
};
\node (detr) [draw=none, fill=none, scale=0.45, xshift=61ex]
{
\tikz
{
\node (cnn) [fill=none, draw=cyan, rounded corners=0.2ex]{CNN};
\node (trans) [fill=none, draw=magenta, rounded corners=0.2ex, xshift=0ex, yshift=-10ex]{Transformer};
\node (box) [fill=none, draw=gray, rounded corners=0.2ex, xshift=-4ex, yshift=-22ex]{Box};
\node (class) [fill=none, draw=gray, rounded corners=0.2ex, xshift=4ex, yshift=-22ex]{Class};
\draw [->, very thick] (cnn) -- (trans);
\draw [->, very thick] ($(trans.south) - (3ex, 0ex)$) -- ($(box.north) + (1ex, 0ex)$);
\draw [->, very thick] ($(trans.south) + (3ex, 0ex)$) -- ($(class.north) - (1ex, 0ex)$);
}
};
\node (detr) [draw=none, fill=none, scale=0.45, xshift=92ex]
{
\tikz
{
\node (cnn) [fill=none, draw=cyan, rounded corners=0.2ex]{CNN};
\node (trans) [fill=none, draw=magenta, rounded corners=0.2ex, xshift=0ex, yshift=-10ex]{LOR};
\node (box) [fill=none, draw=gray, rounded corners=0.2ex, xshift=-4ex, yshift=-22ex]{Box};
\node (class) [fill=none, draw=gray, rounded corners=0.2ex, xshift=4ex, yshift=-22ex]{Class};
\draw [->, very thick] (cnn) -- (trans);
\draw [->, very thick] (trans) -| (box);
\draw [->, very thick] (trans) -| (class);
}
};
\node(a) [xshift=-4.25ex, yshift=-5.3ex]{\scriptsize (a)};
\node(a) [xshift=10.25ex, yshift=-5.3ex]{\scriptsize (b)};
\node(a) [xshift=22.75ex, yshift=-5.3ex]{\scriptsize (c)};
\node(a) [xshift=36.75ex, yshift=-5.3ex]{\scriptsize (d)};
\end{tikzpicture}
\caption{(a) Faster-RCNN, (b) SSD, (c) DETR, and (d) FFD.}
\label{fig:detectors}
\vspace{-4ex}
\end{figure}
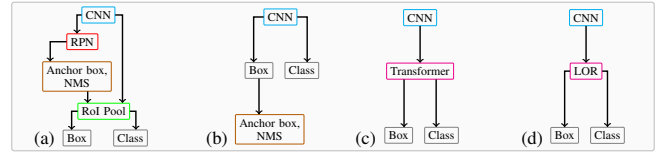
Precisely, we aim to eliminate anchor boxes, NMS and multi-scale detection from a detector. The CNN-only detectors are architecturally simple, and converge faster but have complex training and testing steps, while DETR has simplified training and testing phases but is complex and converges slower \cite{dabdetr}. Moreover, they are configured for large datasets \cite{mscoco} consisting of objects diverse in sizes, aspect ratio, and appearance, leaving room to incorporate task-centric observations when designing a detector. For instance, we target apple-like fruit which is quite small, and hence, efficient detection of small objects can be the main focus.
\par
Since backbone is common among CNN detectors and DETR, with only differences in the prediction head strategy, we revisit both the designs, and re-innovate the detection head. This results in \textit{FFD}, a single-staged, free of RPN, NMS or anchor-box detector having a simplified training and testing phase. Fig.~\ref{fig:detectors} differentiates FFD architecture from the mainstream representative detectors.
\input{figures/ffd}
\subsection{Backbone}
\label{sec:backbone}
A large portion of the runtime is contributed by the backbone, therefore we choose VGG \cite{vgg} network due to its plain structure and lower latency. We enhance it with BatchNorm \cite{bn} for faster convergence and better generalization. It is five staged with $\{2,2,3,3,4\}$ layers and $\{16,32,64,128,256\}$ neurons per stage, each operating at a stride of $2$, and the final one producing a tensor $T_f \in \mathbb{R}^{C \times H_o \times W_o}$, where $C=256$, $H_o=\frac{H}{32}$, and $W_o=\frac{W_o}{32}$, $H, W$ are the image height and width.
\par
As small objects lose their identity in low-resolution feature maps, multi-scale detection \cite{fpn} is employed. However, we aim to detect them only from low-resolution map $T_f$ to reduce computational complexity (Sec.~\ref{sec:rel_cnnobj}). To achieve that, we propose a latent object representation (\texttt{LOR}) module that is motivated by the query-key-value paradigm of \cite{detr} but is free of transformer attention mechanism and is fully convolutional.
\par
\textit{Note:} Backbone can be chosen to be any other network depending on the difficulty of a dataset and desired accuracy.  
\subsection{Latent Object Representation (\texttt{LOR}) }
Here we refer the reader to DETR concepts \cite{detr} to better understand the upcoming text. DETR produces a fixed number of queries, each representing an object. The queries are initialized via embeddings \cite{detr} or anchor-boxes \cite{dabdetr}, and iteratively refined via compute-intensive self-attention and cross-attention of transformer encoder and decoder blocks. Regardless of the query design (embedding or anchor-box), they are not generated from the backbone in any of \cite{detr, dabdetr}. 
\par
On the contrary, we propose to generate them via \texttt{LOR} module (Fig.~\ref{fig:ffd}) directly from the backbone output in a computationally efficient manner, without transformers. This results in an extremely simplified detection pipeline which is also free of post-processing. To the best of our knowledge, this query design is novel and FFD is the first to utilize it.
\par
The \texttt{LOR} module can be divided into two parts: query transformation (\texttt{QT}), and cross-channel global context (\texttt{CCGC}).% We describe each of them in detail below.
\subsubsection{Query Transformation \texttt{QT}}
In this step, the input tensor ($T_i$) to the \texttt{LOR} module is passed through a $1 \times 1$ convolution whose output is added to the input (residual connection \cite{resnet}). \texttt{QT} essentially adds non-linearity to the input queries and results in a tensor $T_{qt}$, denoted as below:
\begin{equation}
T_{qt} = \mathrm{ReLU}(\mathcal{F}_{qt}(T_i) + T_i),~~~\mathcal{F}_{qt} \equiv \mathrm{Conv_{1\times1}}
\end{equation}
\subsubsection{Cross Channel Global Context (\texttt{CCGC})}
The output of the backbone ($T_f$) is devoid of large spatial context due to the shallow backbone that limits its receptive field. However, the role of contextual information in detection and segmentation is crucial \cite{pspnet}. Although there are many ways \cite{pspnet} to do so, we devise a simple and compute efficient strategy \texttt{CCGC}.
\par
In this strategy, we pass the input through global pooling, producing a $1$D tensor $\mathbf{z} \in \mathbb{R}^{C}$ whose $i^{th}$ channel is given by:
\begin{equation}
z_i = \frac{1}{H_o \times W_o} \sum_{h \in H_o, w \in W_o} T_{qt}(h,w)
\end{equation}
where, $H_o, W_o$ are the height, and width of the input tensor.
\par
At this point, elements of $\mathbf{z}$ carry global context but lack cross-channel context. Thus to embed the cross-channel context, $\mathbf{z}$ is transformed via two sequentially connected convolution layers which intertwine the content of $z_i$'s; the first layer expands the input channels by a factor $r$ ($\mathcal{F}_e$) while the other squeezes them by the same factor ($\mathcal{F}_s$), denoted as:
\begin{equation}
\mathcal{F}_e \equiv \mathrm{ReLU(Conv_{1\times1})},~~~\mathcal{F}_s \equiv \mathrm{\sigma(Conv_{1\times1})}
\end{equation}
%
%%
%\begin{equation}
%\mathcal{F}_s \equiv \mathrm{\sigma(Conv_{1\times1})},~~ 
%\end{equation}
%%
where, $\sigma(\cdot)$ stands for Sigmoidal activation.
\par
A similar structure with additional operations is employed in \cite{senet} but is intended to improve CNN's accuracy. On the contrary, our use is entirely different i.e. aggregating global information in a simplified possible manner. 
\par
The resulting tensor is now broadcast multiplied \cite{pytorch} with $T_{qt}$ which weights $T_{qt}$ information depending on the global context. Summarily, \texttt{CCGC} adds non-linearity to $\mathbf{z}$ which is propagated to $T_q$ by amplifying salient information in $T_{qt}$ through broadcast multiplication. \texttt{CCGC} can be written as:
\begin{equation}
\mathcal{F}_{\texttt{ccgc}} \equiv \mathrm{\mathcal{F}_s(\mathcal{F}_e(z))},
\end{equation}
\subsubsection*{Overall Flow}
\texttt{LOR} module takes input the tensor $T_f$ which is operated upon by a $1\times 1$ convolution, producing a tensor $T_g$ of channels $dN_g$, where $d$ is query dimension, and $N_g$ queries exist per spatial location $\in \mathbb{R}^{H_o \times W_o}$ of $T_f$. 
\par
Now \texttt{QT} and \texttt{CCGC} modules are used in parallel and repeated three times to learn better data representation, while still having access to a wider spatial context. Adding more of such modules increases parameters but does not add to accuracy, because the backbone is still fixed. We perform repetition only three times to meet our runtime requirements, however, they are flexible enough to be adjusted. The overall flow of \texttt{LOR} is shown in Fig.~\ref{fig:ffd} and is summarized as follows.
\begin{equation}
\mathcal{F}_{\texttt{LOR}} \equiv \bigodot^{N}_{i=1} T_{qt} \ast \mathcal{F}_{\texttt{ccgc}}
\end{equation}
where, $\odot$ is function-of-function, $\ast$ is broadcast multiplication.
\subsection{Delineation}
The output of \texttt{LOR} is now collapsed spatially, resulting in \textit{query matrix} $T_q \in  \mathbb{R}^{d \times(N_gH_oW_o)}$, whose each row denotes a query $\mathbf{q}$ that represents an object detectable in the image.
\par
In \texttt{LOR}, queries in the form of learnable embeddings \cite{detr} or anchor-box \cite{dndetr, dabdetr} are not needed, instead they are directly generated from the backbone output. This is the major novelty of the \texttt{LOR} module, leading to a simplified structure, high accuracy without needing post-processing, and faster speeds.
\subsection{Prediction}
The tensor $T_q$ is forwarded to two Feed Forward Networks (FFN) which are a stack of $1\times 1$ convolutions followed by ReLU \cite{detr}; One for Classification (FFN$_c$) having one layer, and one for box regression (FFN$_b$), having three layers.
\subsection{Query Assignment}
DETR predicts w.r.t. the image origin ($0,0$), whereas \cite{dabdetr, dndetr} predicts w.r.t. the learned anchors. It limits the total number of detectable objects in the image, regardless of the image resolution. To handle that, we propose to generate $N_g$ queries per spatial location of $T_f$, and each such location refers to a non-overlapping tile of the input image following \cite{vit}. With this strategy, each set of $N_g$ queries in $T_q$ corresponds to all the objects whose center lies in a particular tile (Fig.~\ref{fig:qa}). It is the uniqueness of FFD queries in contrast to DETR \cite{detr}.
\input{figures/query_assignment}
\subsection{Tiled Hungarian Matching}
Ground-truth matching is a crucial step to train an object detector which is performed via region proposal matching \cite{fasterrcnn} and box mining \cite{ssd}. It is full of hyperparameters and is a complicated process (Sec.~\ref{sec:rel_cnnobj}). To avoid that, we use bipartite matching using Hungarian algorithm inspired by \cite{detr} for assigning a ground-truth exactly one prediction, but performing it over tiles instead of the whole image space \cite{detr}.
\par
As mentioned previously that in our case, all of the $N_g$ predictions for each tile are made w.r.t. the top-left corner of that corresponding tile, therefore to match a ground-truth box with a prediction, the prediction is denormalized via Eq.~\ref{eq:denorm} and a cost is computed using $L_{match}$ (discussed next). This is done for each ground-truth box whose center falls into that tile, resulting in a cost matrix $\mathcal{C} \in \mathbb{R}^{G \times N_g}$, where $G$ denotes the number of ground truth boxes falling into a tile. Now, Hungarian matching is performed over $\mathcal{C}$ which assigns a ground-truth box exactly to one prediction $\in [0,N_g)$. This process is performed for all the tiles over the image.
\begin{equation}
\hat{b} = \{\nicefrac{(b_{cx} - g_x)}{g_w}, \nicefrac{(b_{cy} - g_y)}{g_h}, \log(\nicefrac{b_{w}}{W}), \log(\nicefrac{b_{h}}{H}) \}
\end{equation}
\begin{equation}
b = \{\hat{b}_{cx} g_w + g_x,~~\hat{b}_{cy} g_h + g_y,~~\exp(\hat{b}_{w}) W,~~\exp(\hat{b}_{h}) H \}
\label{eq:denorm}
\vspace{-0.4ex}
\end{equation}
where, $\hat{b}$ is the prediction, $b$ is denormalized box, $W, H$ are the image width and height, and ($g_x, g_y$) is the top-left corner of the tile , and $g_w, g_h$ are tile width and height respectively.
\par
Tiled Hungarian matching is different from \cite{detr}, \cite{dabdetr}. \textit{First}, since not all tiles are occupied, it prevents most tiles from performing the matching process, and \textit{Second} not many objects are present in a tile. Together it drastically reduces matching complexity. The claims are verified in Table~\ref{tab:infer}.
%
%
%
%\vspace{-0.15ex}
%
\subsection{Objective Function}
The objective function is a weighted combination of a classification loss (Cross-Entropy) and a box regression (Smooth-$L1$) loss \cite{fasterrcnn}, formulated as below:
\begin{equation}
\mathcal{L}_c = -\log(p)
\end{equation}
\begin{equation}
\mathcal{L}_b = 
\begin{cases}
0.5 (b-\hat{p})^2 / \beta, ~~~  \text{if} (b-\hat{b}) < 1 \\
(b-\hat{b}) - 0.5 \beta, ~~~  \text{otherwise}
\end{cases}
\end{equation}
\begin{equation}
\mathcal{L} = \mathcal{L}_{c} + \lambda  \mathcal{L}_{b}
\label{eq:loss}
\end{equation}
where, $\lambda$ is the loss weight which is set to $1$, and balances the contribution of both losses. $p$ and $b$ are the class logits and box predictions, respectively. The overall objective $\mathcal{L}$ also serves as $\mathcal{L}_{match}$, which is used in the matching process.
\subsection{Inference}
Our inference strategy is free of any post-processing unlike popular approaches \cite{fasterrcnn, ssd, fcos} due to the set predictions and one-to-one matching in contrast to the one-to-many assignment of  \cite{fasterrcnn, ssd, fcos} (discussed previously), resulting in the elimination of NMS entirely, and reduced CPU/GPU occupancy of FFD.
\par
\begin{figure}[t]
\centering

\FPeval{\scal}{0.5}

\colorlet{tensorclr}{white!90!blue}
\colorlet{tensordclr}{white!50!black}

\colorlet{maxclr}{white!80!black}
\colorlet{convclr}{white!80!green}
\colorlet{reluclr}{white!90!black}
\colorlet{eltclr}{white!80!blue}
\colorlet{softmaxclr}{white!80!blue}

\colorlet{drawclr}{white!99!black}
\colorlet{posencclr}{white!50!orange}

\begin{tikzpicture}
\node (outer) [scale=1.0]
{
\tikz{
\node (boundary) [draw=white!90!gray,  fill=white!98!black,rounded corners=0.2ex, minimum width=51ex, minimum height=23ex, xshift=22ex,yshift=-4ex]{};
\node (backbone) [draw=gray, fill=white!50!cyan, rounded corners=0.25ex, minimum width=6ex, minimum height=2.5ex, xshift = 0ex, scale=0.8]{\scriptsize BackBone};
\node (ffn) [draw=gray, fill=white!70!yellow, rounded corners=0.25ex, minimum width=6ex, minimum height=2.5ex, xshift = 45ex, scale=0.8]{\scriptsize FFN};
\node (ftfd) [xshift=23ex, yshift=4.5ex]
{
\tikz{
\node (boundary) [draw=gray, dashed, dash pattern=on 0.8ex off 0.3ex, rounded corners=0.2ex, minimum width=40ex, minimum height=5.1ex, xshift=7.95ex, yshift=0.95ex]{};
\node (lqg) [draw=gray, fill=white!70!violet, rounded corners=0.25ex, minimum width=6ex, minimum height=3ex, xshift = 1.1ex, scale=0.8]{\scriptsize Latent Object Representation (Object Queries) };
\node (qr) [draw=gray, fill=white!50!orange, rounded corners=0.25ex, minimum width=13ex, minimum height=3ex, xshift = 22ex, scale=0.8]{\scriptsize Dilineation};
\node (text) [xshift=8.25ex, yshift=2.25ex]{\scriptsize FFD};
\draw [->] (lqg.east) -- (qr.west);
}
};
\node (detr) [xshift=22ex, yshift=-8.5ex]
{
\tikz{
\node (boundary) [draw=gray, dashed, dash pattern=on 0.8ex off 0.3ex, rounded corners=0.2ex, minimum width=50ex, minimum height=13ex, xshift=0ex, yshift=0ex]{};
\node (selfatt) [draw=gray, fill=white!70!violet, rounded corners=0.25ex, minimum width=6ex, minimum height=3ex, xshift = -12ex, yshift=2.1ex, scale=0.8]{\scriptsize Multi-head Self Attention};
\node (ffne) [draw=gray, fill=white!50!blue, rounded corners=0.25ex, minimum width=6ex, minimum height=3ex, xshift = 0ex, yshift=2.1ex, scale=0.8]{\scriptsize FFN};
\node (text) [xshift=-8ex, yshift=4.5ex]{\scriptsize Transformer Encoder};
\node (enctimes) [xshift=1.5ex, yshift=4.5ex]{\scriptsize $\times 6$};
\node (encb) [draw=gray, dashed, dash pattern=on 0.8ex off 0.3ex, rounded corners=0.2ex, minimum width=22.5ex, minimum height=5.25ex, xshift=-8.3ex, yshift=3.1ex]{};
\node (qr) [draw=gray, fill=white!50!orange, rounded corners=0.25ex, minimum width=6ex, minimum height=3ex, xshift = -19.5ex, yshift=-4.2ex, scale=0.8]{\scriptsize Object Queries)};
\node (selfatt1) [draw=gray, fill=white!70!violet, rounded corners=0.25ex, minimum width=6ex, minimum height=3ex, xshift = -6.6ex, yshift=-4.2ex, scale=0.8]{\scriptsize Multi-head Self Attention};
\node (crossatt) [draw=gray, fill=white!60!magenta, rounded corners=0.25ex, minimum width=6ex, minimum height=3ex, xshift = 9.25ex, yshift=-4.2ex, scale=0.8]{\scriptsize Multi-head Cross Attention};
\node (ffnd) [draw=gray, fill=white!50!blue, rounded corners=0.25ex, minimum width=6ex, minimum height=3ex, xshift = 20.25ex, yshift=-4.2ex, scale=0.8]{\scriptsize FFN};
\node (textd) [xshift=4ex, yshift=-1.7ex]{\scriptsize Transformer Decoder};
\node (dectimes) [xshift=21.5ex, yshift=-1.7ex]{\scriptsize $\times 6$};
\node (decb) [draw=gray, dashed, dash pattern=on 0.8ex off 0.3ex, rounded corners=0.2ex, minimum width=37.5ex, minimum height=5.25ex, xshift=4.5ex, yshift=-3.2ex]{};
\node (text) [xshift=18.25ex, yshift=5.15ex]{\scriptsize DETR};
\draw [->] (qr.east) -- (selfatt1.west);
\draw [->] (selfatt1.east) -- (crossatt.west);
\draw [->] (crossatt.east) -- (ffnd.west);
\draw [->] (selfatt.east) -- (ffne.west);
\draw [->] (ffne.east) -| ($(crossatt.north)+(3ex,0ex)$);
}
};
\draw [->] (backbone.north) |- ($(ftfd.west)+(1.75ex,-0.9ex)$);
\draw [->] ($(ftfd.east)-(1.85ex,0.9ex)$) -| (ffn.north);
\draw [->] (backbone.south) |- ($(detr.west)+(6.8ex,2.1ex)$);
\draw [->] ($(detr.east)-(3.2ex,4.2ex)$) -| ($(ffn.south)+(1ex,0ex)$);
}
};
\end{tikzpicture}
\vspace{-0.5ex}
\caption{Differences between FFD and DETR-like methods.}
\vspace{-3ex}
\label{fig:diffdetr}
\end{figure}
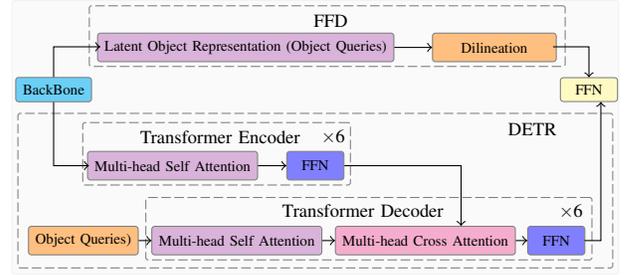

Further, in FFD, all the predictions are made w.r.t. the top-left corner of a tile, therefore they are denormalized by using Eq.~\ref{eq:denorm} before the final use. The overall information flow of FFD is depicted in Fig.~\ref{fig:ffd}. Also, we have shown the difference between FFD and DETR \cite{detr} in Fig.~\ref{fig:diffdetr}.
\section{Occlusion Aware Scene Synthesis}
\label{sec:occ}
CNN-based algorithms are sensitive to the amount of a dataset, if it is limited, the network may overfit and perform poorly. In our context, one image consists of several instances of apples which turns manual annotation of images an exhaustive and time-consuming task. Therefore, collecting many images and labelling them become a key challenge.
\par
Hence we contribute by adapting occlusion-aware scene synthesis from our previous work \cite{semi}. The original approach generates realistic cluttered scenes from isolated object images when it is difficult to label real cluttered images. In this technique, an image called \texttt{base\_image} is picked randomly from the dataset and is divided into a grid of $K\times K$. Now, another image from the dataset is chosen randomly, and pixels corresponding to an object instance in this image are pasted onto the grid center of one of the grids in the \texttt{base\_image}. This procedure is repeated $K\times K$ grid locations. $K$ is randomly chosen from $3\times3$, $4\times4$, $5\times5$ to simulate low, mid, and high clutter. Finally, instances below a visibility threshold ($25\%$) are filtered out. See \cite{semi} for more details.
\par
The data is then used for the task of semantic segmentation where it doesn't matter even if the object is visible by $25\%$. However, in object detection, as our objects are already too small, this approach generates cluttered images with too many overlapping and meaningless instances (Fig.~\ref{fig:synold}, ~\ref{fig:synoldrect}).
\input{figures/synthetic_clutter}
\input{figures/dataset_images}
%%%
To adapt this approach to our use case, we make two changes. \textit{First}, there is no notion of grids, instead maximum number of instances per image $N_{max}$ is defined i.e. we use random locations instead of fixed grids, and \textit{Second}, we put a constraint that none of the boxes overlap with each other.
\par
In order to generate a synthetic scene based on the above changes, we begin by randomly selecting an image $\mathrm{I}^0_s$ without any fruits (\texttt{base\_image}). Then we randomly choose a number $N_i \in [0,N_{max})$ which defines the number of instances the resulting synthesised image $\mathrm{I}_s$ will contain. Now, we randomly pick an image carrying fruit instances and its corresponding mask ground truth. With the help of the mask, the number of instances in this image is computed, and one of the instances is selected randomly to be transferred to the \texttt{base\_image}. Now, a random location in $\mathrm{I}^0_s$ is sampled, and before pasting the contents of the selected instance, it is ensured that the bounding box of this instance doesn't overlap with any of the instances already pasted during this process if placed at the sampled location. This procedure is repeated $N_i$ times and can be summarized as:
%
%%
%\begin{equation}
%\mathrm{I}^{t}_s = \mathrm{M}^{t}_{xy} * \mathrm{I}^{t-1}_s + \mathrm{P}^{t}_{xy},~~t \in (0, N_i]
%\end{equation}
%%
%%
%\begin{equation*}
%s.t.~~ \mathrm{R}^{t}_{xy} \cap \mathrm{R}^{t-1}_{xy} = \phi~~~\forall~t-1 \in (0, t)
%\end{equation*}
%
%
\begin{align}
\mathrm{I}^{t}_s = \mathrm{M}^{t}_{xy} * \mathrm{I}^{t-1}_s + \mathrm{P}^{t}_{xy},~~t \in (0, N_i] \\
s.t.~~ \mathrm{R}^{t}_{xy} \cap \mathrm{R}^{t-1}_{xy} = \phi~~~\forall~t-1 \in (0, t)
\end{align}
where, $\mathrm{P}^{t}_{xy}, \mathrm{M}^{t}_{xy},  \mathrm{R}^{t}_{xy}$ refers to the patch, its mask, and its bounding box respectively. We set $N_{max} = 100$.
\par
Fig.~\ref{fig:newclutter} shows synthetic scenes generated by the original \cite{semi} and our improved method. It can be noticed that the scenes generated by our method make much more sense in terms of visual quality as well as from the training perspective.
%%
%%%
%%
\section{Dataset}
\label{sec:dataset}
Due to the lack of orchards in our vicinity, we build an farming setup (Fig.~\ref{fig:harvesting}). It facilitates round-the-clock testing of aerial grasping without waiting for appropriate weather.
\par
We collect two datasets: (\texttt{i}) $D_t$: fruit hanging over an artificial tree (Fig.~\ref{fig:imtree}), and (\texttt{ii}) $D_h$: fruit hanging over the harvesting region  (Fig.~\ref{fig:imharvest}). We collect $150$ images for each case, both indoor and outdoor. Following \cite{semi}, we also collect $25$ images for each of the trees and the harvesting region without any fruit to serve as the \texttt{base\_image} for generating synthetic scenes. In addition, we collect a few images of real trees with apples manually attached to it. It is done in order to test the robustness and generalization of FFD across scenes. %Our dataset is binary class owing to the fact that a tree/plant can only have one kind of fruit \cite{minneapple}.
\subsection{Labelling Process}
For each image, a mask is generated whose pixels indicate class labels; Background has label $0$ while fruit has label $1$. Box annotations are extracted from the convex hull of the pixels belonging to an instance in the mask. Masks facilitate rotation augmentation since rotating a bounding box annotation does not precisely enclose the rotated object.
\subsection{Synthetic Scenes}
Manual labeling took $5$-$8$ minutes per image of the harvesting region dataset due to a large number of instances, necessitating our scene synthesis technique (Sec.~\ref{sec:occ}). Fig.~\ref{fig:cluttertree} shows a few samples of synthesised scenes. 
\begin{figure}[!t]
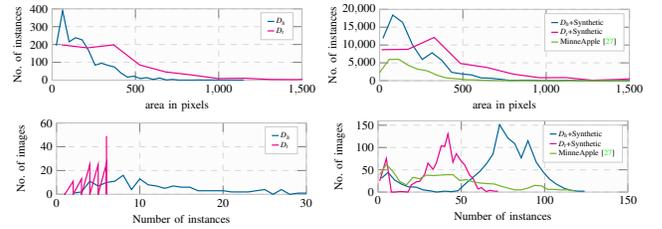

\centering

\begin{tikzpicture}

\FPeval{\xshfta}{0}
\FPeval{\xshftb}{27}
\FPeval{\xshftc}{0}
\FPeval{\xshftd}{27}

\FPeval{\yshfta}{0-0}
\FPeval{\yshftb}{0-0}
\FPeval{\yshftc}{0-9.5}
\FPeval{\yshftd}{0-9.5}

\FPeval{\pltw}{38}
\FPeval{\plth}{18}

\FPeval{\scal}{0.75}

\FPeval{\mrksize}{0.3}

\colorlet{trellisclr}{blue!60!green}
\colorlet{treeclr}{white!0!magenta}
\colorlet{minneappleclr}{brown!60!green}

\colorlet{gridclr}{white!85!black}
\colorlet{dlegendclr}{white!80!black}
\colorlet{axisclr}{white!75!black}
\colorlet{axisbgclr}{white!99!black}

\colorlet{dlegendclr}{white!80!black}
\colorlet{legendclr}{white!100!black}

\FPeval{\lscale}{0.4}
\FPeval{\limscale}{0.2}

\FPeval{\lxshfta}{25.2}
\FPeval{\lyshfta}{4.5}
\FPeval{\lxshftb}{22.95}
\FPeval{\lyshftb}{3.3}
\FPeval{\lxshftc}{25.2}
\FPeval{\lyshftc}{4.5}
\FPeval{\lxshftd}{22.95}
\FPeval{\lyshftd}{3.3}

\FPeval{\titlescale}{0.7}
\FPeval{\titlexshift}{21.0}
\FPeval{\titleyshift}{13.7}

\colorlet{gpudclr}{white!80!black}
\colorlet{gpuclr}{white!90!black}
\colorlet{gputxtclr}{white!0!black}

\FPeval{\labelscale}{0.58}
\FPeval{\ticklabelscale}{0.6}

\FPeval{\xlabelxshifta}{0+14}
\FPeval{\xlabelyshifta}{0+1.5}
\FPeval{\ylabelxshifta}{0+3.9}
\FPeval{\ylabelyshifta}{0+4.0}

\FPeval{\xlabelxshiftb}{0+14}
\FPeval{\xlabelyshiftb}{0+1.5}
\FPeval{\ylabelxshiftb}{0+2.5}
\FPeval{\ylabelyshiftb}{0+4.0}

\FPeval{\xlabelxshiftc}{0+14}
\FPeval{\xlabelyshiftc}{0+1.5}
\FPeval{\ylabelxshiftc}{0+3.9}
\FPeval{\ylabelyshiftc}{0+4.0}

\FPeval{\xlabelxshiftd}{0+14}
\FPeval{\xlabelyshiftd}{0+1.5}
\FPeval{\ylabelxshiftd}{0+3.5}
\FPeval{\ylabelyshiftd}{0+4.0}

\FPeval{\linew}{0.6}
\FPeval{\dashon}{5.0}
\FPeval{\dashoff}{3.0}

\FPeval{\xshfttakeoff}{0-16}
\FPeval{\xshfthover}{0}
\FPeval{\xshftland}{16}
\node (a) [xshift = \xshfta ex, yshift=\yshfta ex, scale=\scal]{
\tikz{
\pgfplotsset{width=\pltw ex, height=\plth ex}
\begin{axis}[
   axis background style={fill=axisbgclr},
    title={},
    xlabel={area in pixels},
    ylabel={No. of instances},
    xmin=0, xmax=1500,
    ymin=0, ymax=400,
     axis line style={axisclr},
    legend image post style={scale =\limscale},
    legend style={at={(\lxshfta ex,\lyshfta ex)},anchor=south, legend columns = 1, draw = {dlegendclr}, fill={legendclr}, nodes={scale=\lscale}},
    ymajorgrids=true, 
    xmajorgrids=true,
    grid style={dashed, gridclr},
    major tick length=1ex,
    x label style={at={(\xlabelxshifta ex, \xlabelyshifta ex)},scale=\labelscale},
    y label style={at={(\ylabelxshifta ex, \ylabelyshifta ex)},scale=\labelscale},
    xticklabel style={scale=\ticklabelscale},
    yticklabel style={scale=\ticklabelscale},
    legend cell align={left},
%    ybar,
%    bar width = 4pt,
]
\FPeval{\opacty}{1.0}
\input{figures/dataset_stats/trellis}
\input{figures/dataset_stats/tree}
  \legend{$D_h$, $D_t$}
\end{axis}
%
% \node [draw=gpudclr,fill=gpuclr,rounded corners=0.2ex, minimum width=10.0ex, xshift=\titlexshift ex, yshift=\titleyshift ex, scale=\titlescale]{\textcolor{gputxtclr}{\textbf{Step-X}}};
}};
\node (a) [xshift = \xshftb ex, yshift=\yshftb ex, scale=\scal]{
\tikz{
\pgfplotsset{width=\pltw ex, height=\plth ex}
\begin{axis}[
   axis background style={fill=axisbgclr},
    title={},
    xlabel={area in pixels},
    ylabel={No. of instances},
    xmin=0, xmax=1500,
    ymin=0, ymax=20000,
     axis line style={axisclr},
    legend image post style={scale =\limscale},
    legend style={at={(\lxshftb ex,\lyshftb ex)},anchor=south, legend columns = 1, draw = {dlegendclr}, fill={legendclr}, nodes={scale=\lscale}},
    ymajorgrids=true, 
    xmajorgrids=true,
    grid style={dashed, gridclr},
    major tick length=1ex,
    x label style={at={(\xlabelxshiftb ex, \xlabelyshiftb ex)},scale=\labelscale},
    y label style={at={(\ylabelxshiftb ex, \ylabelyshiftb ex)},scale=\labelscale},
    xticklabel style={scale=\ticklabelscale},
    yticklabel style={scale=\ticklabelscale},
    legend cell align={left},
    scaled y ticks={base 10:-0}
%    ybar,
%    bar width = 4pt,
]
\FPeval{\opacty}{1.0}
\input{figures/dataset_stats/trellis_and_trellis_synth}
\input{figures/dataset_stats/tree_and_tree_synth}
\input{figures/dataset_stats/minneapple}

  \legend{$D_h$+Synthetic, $D_t$+Synthetic, MinneApple \cite{minneapple}}
\end{axis}
%
% \node [draw=gpudclr,fill=gpuclr,rounded corners=0.2ex, minimum width=10.0ex, xshift=\titlexshift ex, yshift=\titleyshift ex, scale=\titlescale]{\textcolor{gputxtclr}{\textbf{Step-X}}};
}};
\node (c) [xshift = \xshftc ex, yshift=\yshftc ex, scale=\scal]{
\tikz{
\pgfplotsset{width=\pltw ex, height=\plth ex}
\begin{axis}[
   axis background style={fill=axisbgclr},
    title={},
    xlabel={Number of instances},
    ylabel={No. of images},
    xmin=0, xmax=30,
    ymin=0, ymax=60,
     axis line style={axisclr},
    legend image post style={scale =\limscale},
    legend style={at={(\lxshftc ex,\lyshftc ex)},anchor=south, legend columns = 1, draw = {dlegendclr}, fill={legendclr}, nodes={scale=\lscale}},
    ymajorgrids=true, 
    xmajorgrids=true,
    grid style={dashed, gridclr},
    major tick length=1ex,
    x label style={at={(\xlabelxshiftc ex, \xlabelyshiftc ex)},scale=\labelscale},
    y label style={at={(\ylabelxshiftc ex, \ylabelyshiftc ex)},scale=\labelscale},
    xticklabel style={scale=\ticklabelscale},
    yticklabel style={scale=\ticklabelscale},
    legend cell align={left},
%    ybar,
%    bar width = 4pt,
]
\FPeval{\opacty}{1.0}
\input{figures/dataset_stats/trellis_n_instance}
\input{figures/dataset_stats/tree_n_instance}
  \legend{$D_h$, $D_t$}
\end{axis}
%
% \node [draw=gpudclr,fill=gpuclr,rounded corners=0.2ex, minimum width=10.0ex, xshift=\titlexshift ex, yshift=\titleyshift ex, scale=\titlescale]{\textcolor{gputxtclr}{\textbf{Step-X}}};
}};
\node (d) [xshift = \xshftd ex, yshift=\yshftd ex, scale=\scal]{
\tikz{
\pgfplotsset{width=\pltw ex, height=\plth ex}
\begin{axis}[
   axis background style={fill=axisbgclr},
    title={},
    xlabel={Number of instances},
    ylabel={No. of images},
    xmin=0, xmax=150,
    ymin=0, ymax=160,
     axis line style={axisclr},
    legend image post style={scale =\limscale},
    legend style={at={(\lxshftd ex,\lyshftd ex)},anchor=south, legend columns = 1, draw = {dlegendclr}, fill={legendclr}, nodes={scale=\lscale}},
    ymajorgrids=true, 
    xmajorgrids=true,
    grid style={dashed, gridclr},
    major tick length=1ex,
    x label style={at={(\xlabelxshiftd ex, \xlabelyshiftd ex)},scale=\labelscale},
    y label style={at={(\ylabelxshiftd ex, \ylabelyshiftd ex)},scale=\labelscale},
    xticklabel style={scale=\ticklabelscale},
    yticklabel style={scale=\ticklabelscale},
    legend cell align={left},
    scaled y ticks={base 10:-0}
%    ybar,
%    bar width = 4pt,
]
\FPeval{\opacty}{1.0}
\input{figures/dataset_stats/trellis_and_trellis_synth_n_instance}
\input{figures/dataset_stats/tree_and_tree_synth_n_instance}
\input{figures/dataset_stats/minneapple_n_instance}

  \legend{$D_h$+Synthetic, $D_t$+Synthetic, MinneApple \cite{minneapple}}
\end{axis}
%
% \node [draw=gpudclr,fill=gpuclr,rounded corners=0.2ex, minimum width=10.0ex, xshift=\titlexshift ex, yshift=\titleyshift ex, scale=\titlescale]{\textcolor{gputxtclr}{\textbf{Step-X}}};
}};

\end{tikzpicture}
\vspace{-1.5ex}
\caption{Comparison of our dataset and MinneApple \cite{minneapple} benchmark.}
\label{fig:datasetstats}
\vspace{-1ex}
\end{figure}
%
%
%
%%

%

%
%%%%%
%
%%
\begin{table}[!t]

\centering
\caption{Dataset statistics comparison.}
\label{tab:datasetstats}

\arrayrulecolor{white!60!black}
\scriptsize
%\footnotesize
%\tiny

\setlength{\tabcolsep}{6.5pt}

\vspace{-0.5ex}

\begin{tabular}{l c c}
\midrule

Dataset & \#Average Size (pixels) & \#Average instance per image \\ \midrule

$D_h$ &  $13\times 13$ &  $13$ \\ 
$D_t$  &  $20 \times 20$ &  $ 6$ \\ 
$D_h$ + Synthetic  &  $14\times 14$ &  $\mathbf{74}$ \\ 
$D_t$ + Synthetic  &  $20 \times 20 $ &  $40$ \\ 
MinneApple \cite{minneapple}  &  $13 \times 13$ &  $42$ \\ 

  \bottomrule

\end{tabular}
\vspace{-3.5ex}
\end{table}
\subsection{Comparison With Existing Benchmarks}
Fig.~\ref{fig:datasetstats} compares our dataset with the existing benchmark. The most closely related is the recent MinneApple \cite{minneapple} dataset consisting of images from apple orchards. It offers bounding boxes and masks for each instance. We see that, our dataset has many instances which are very small that are challenging for detectors. This is a unique aspect of our dataset.
\par
Moreover, the MinneApple benchmark has two major issues. \textit{First}, masks for many instances are missing, and \textit{Second}, it also consists of several fruit instances which lie on the ground. The ground instances are not annotated, instead, only the ones on the tree are annotated. It results in an unfair evaluation because the ground instances resemble the ones on the tree and are detected by the detector. On the other hand, our dataset is free of such issues. See video.
\par
Despite the advantages, our dataset has its own limitations, e.g. it does not include many occluded instances as compared to MinneApple, and it has less scenic diversity. Nonetheless, our dataset can be used for extensive verification during the initial development phase of new detectors, and later MinneApple-like datasets can be used.
\input{figures/detection_results}
\section{Experiments}
\label{sec:exp}

\begin{table}[t]

\centering

\caption{Cross-dataset performance of FFD.}
\label{tab:detector}

\arrayrulecolor{white!60!black}
\scriptsize
%\footnotesize
%\tiny

\setlength{\tabcolsep}{7.6pt}

\vspace{-0.5ex}
\begin{tabular}{c c c c c c c}
\toprule

\multicolumn{1}{c}{Exp} & \multicolumn{1}{c}{Train dataset} & \multicolumn{1}{c}{Test dataset} & AP &  AP$_{S}$ &  AP$_{M}$ &  AP$_{L}$  \\ \cmidrule{1-7}

%\multicolumn{1}{c|}{\multirow{2}{*}{\texttt{Exp$1$}}} & \multirow{2}{*}{Tree} & Tree & $28.7$ & $59.0$ & $23.7$  & $24.4$  & $28.8$  & $35.0$ \\ 
\multicolumn{1}{c}{\multirow{2}{*}{\texttt{E$1$}}} & \multirow{2}{*}{$D_t$} & $D_t$ & $51.1$  & $28.6$  & $51.7$  & $65.7$ \\ 
\multicolumn{1}{c}{}  & & $D_h$ & $36.8$  & $17.2$  & $42.2$  & $-$ \\ \midrule
 % & Outdoor & $95.1$ & $100$  & $-$  & $-$  & $-$  & $-$ \\ \midrule

\multicolumn{1}{c}{\multirow{2}{*}{\texttt{E$2$}}} & \multirow{2}{*}{$D_h$} & $D_t$ & $30.2$  & $23.5$  & $34.6$  & $19.1$ \\ 
\multicolumn{1}{c}{}  & & $D_h$ & $46.6$  & $31.0$ & $52.1$ & $-$ \\ \midrule
  %& Outdoor & $82.8$ & $100$  & $-$  & $-$  & $-$  & $-$ \\ 

\multicolumn{1}{c}{\multirow{2}{*}{\texttt{E$3$}}} & \multirow{2}{*}{\makecell{$D_t$ + $D_h$}} & $D_t$ & $53.9$ & $23.1$  & $55.2$  & $67.4$ \\ 
\multicolumn{1}{c}{} &  & $D_h$ & $49.1$ & $31.4$  & $54.2$  & $-$ \\ 
%  & Outdoor & $83.1$ & $100$  & $-$  & $-$  & $-$  & $-$ \\ 

\bottomrule

\end{tabular}
\vspace{-3.25ex}
\end{table}
\subsection{Training Hyperparameters}
We set $\texttt{base\_lr}=0.001$, and use CosineAnnealing scheduler \cite{cosineanneal} with $\texttt{weight\_decay}=0.0001$, and \texttt{ADAM} optimizer with $\beta_1=0.90, \beta_2=0.99$ for $1000$ epochs.
\subsection{Comprehensive Data Augmentation}
We use runtime augmentation \cite{semi} i.e. hue, saturation, brightness, and contrast perturbation with a likelihood of $0.4$, random rotation in $[-10^{o},10^{o}]$, random translation in $[-50, 50]$ pixels, mirror, and scale. This prevents overfitting by accounting for lighting, and geometric transformations.
\subsection{Training Policy}
We split the datasets into a train-test ratio of $2:1$, while the outdoor images are used only for evaluation. We perform three experiments; \textit{First}, \texttt{E$1$}: Train on $D_t$ and test all, \textit{Second}, \texttt{E$2$}: Train on $D_h$ and test all, and \textit{Third}, \texttt{E$3$}: Train on both $D_t$ and $D_h$ and test all. The resolution is set to $320\times 256$.
\subsection{Quantitative Evaluation}
We report Average-Precision (AP) \cite{fasterrcnn} to evaluate FFD, and AP$_{S}$, AP$_{M}$, AP$_{L}$ for instances having different area (in pixels) i.e. small ($[0, 10^2]$), medium ($(10^2, 30^2]$), and large ($>30^2$).
\par
Table~\ref{tab:detector} shows the analysis of the three experiments. It can be seen that FFD performs with sufficiently high AP score, also verifiable via qualitative evaluations, discussed next.
\par
It is interesting to note that cross-dataset testing has inferior performance when only one dataset is used for training (\texttt{E$1$} or \texttt{E$2$}). As per our observations, FFD was able to detect all the instances on the cross-dataset, but AP dropped because of the misclassification of certain fruit-like spots in the images. It happened due to the lack of scenic diversity.
\par
Furthermore, FFD has slightly higher AP in \texttt{E$1$} relative to \texttt{E$2$}. This indicates the challenging nature of harvesting region dataset due to the presence of small instances.
\par
\texttt{E$3$} shows that using both datasets improves the accuracy for each of them, owing to the increased image diversity.
\subsubsection{Qualitative Results}
Fig.~\ref{fig:qualitative} shows a few detection samples from the test-set, and also detections on outdoor images which none of the experiments used for training. From the detection quality, it can readily be verified that the detections have a very high overlap with the ground-truth boxes, which is a most required attribute for robotic harvesting autonomy. This facilitates accurate centroid calculation using depth information, a crucial step for performing a robust visual servoing and grasping operation using UAV.
\begin{table}[t]

\centering

\caption{Detection Performance on our dataset. FFD has very high detection performance while still being faster than all the baselines.}
\label{tab:train}

\arrayrulecolor{white!60!black}
\scriptsize
%\footnotesize
%\tiny

\setlength{\tabcolsep}{14.2pt}

\vspace{-0.5ex}
\begin{tabular}{l | c c c c c}
\toprule

\multicolumn{1}{c|}{Detector} & AP  &  AP$_{S}$ &  AP$_{M}$ \\ \midrule

SSD $@$multi-scale \cite{ssd} & $38.0$  & $20.1$  & $39.1$ \\ 
DETR $@$multi-scale \cite{detr} & $40.2$  & $24.9$ & $43.0$ \\ 
FCOS $@$multi-scale \cite{fcos} & $42.1$  & $26.8$  & $46.9$ \\ 
Faster-RCNN $@$multi-scale \cite{fasterrcnn} & $45.9$ & $28.7$  & 
$51.5$ \\ 
YOLO-v$8$ $@$multi-scale \cite{yolov8} & $46.3$ & $29.2$  & 
$51.5$ \\ 
YOLO-v$8$ $@$single-scale \cite{yolov8} & $35.2$  & $23.1$ & $41.4$ \\ \midrule 
FFD $@$single-scale & $\mathbf{46.6}$ & $\mathbf{31.2}$  & $\mathbf{52.1}$  \\ 

\bottomrule

\end{tabular}
\vspace{-4.0ex}
\end{table}
\subsection{Detection Performance Against Exiting Detectors}
We compare FFD with popular and well-established detectors by customizing them for our dataset. This task itself is challenging because each detector has its source code implemented differently in different frameworks. This raises the difficulty level to analyze each of them. Hence the baselines are selected such that it covers almost all the varieties of the detectors, i.e. multi-stage \cite{fasterrcnn}, single stage \cite{ssd, fcos} and transformer-based \cite{detr} to minimize the retraining efforts. We leave DETR's successor \cite{dabdetr} due to its highly complex and resource-hungry training.%, and \cite{bochkovskiy2020yolov4} due to its backbone-only modifications and traditional detection process.
\subsubsection{Detection on Our Dataset}
Table~\ref{tab:train} shows the corresponding analysis. The baselines are trained with our backbone (Sec.~\ref{sec:backbone}), on the harvesting region dataset due to its higher difficulty. From the table, we can see that FFD is as accurate as the most complex detector Faster-RCNN \cite{fasterrcnn} and the latest YOLO-v$8$, including small objects. But FFD outperforms them in single-scale comparison, i.e. when Faster-RCNN and YOLO are trained for single-scale detection only similar to FFD. It even performs better than the transformer-based DETR \cite{detr}, as DETR converges slowly; however, it can be trained longer to achieve comparable accuracy.
\par
FFD earns this upper hand only because of the \texttt{LOR} module, precisely due to the query generation from the feature map and the prediction strategy, which is the main novelty of FFD, along with its unique training scheme. 
\par
\textit{Note:} Accuracy can be improved by changing the backbone. We fixed the backbone and kept sufficiently large epochs to meet our speed and resource requirements.
\subsubsection{Detection on MinneApple Benchmark \cite{minneapple}}
We also conduct experiments on the recent MinneApple benchmark for apple detection (See Table~\ref{tab:det_minneapple}). Noticeably, with ResNet-$50$ backbone and different tile sizes, FFD achieves similar detection scores at single-scale detection while being considerably faster. Most importantly, Faster-RCNN performs multi-scale detection \cite{minneapple}, which is still slower than FFD. Moreover, the latest YOLO-v$8$ performs worse in single-scale settings but is comparable to FFD in multi-scale. This shows the uniqueness of FFD that despite being single-scale, it outperforms multi-scale methods.
\par
Since MinneApple is a challenging dataset, it needs a bigger backbone. However, we also tried our smaller backbone, which obtains a lower AP. It is evident due to its fewer parameters, i.e. only $3$M vs $25$M of ResNet-$50$. %However, it can be scaled easily depending on the requirement. While in terms of runtime, FFD is exceptionally fast. 
\begin{table}[!t]

\centering
\caption{Evaluation on MinneApple \cite{minneapple}. `$^\star$' denotes our results. All networks are trained with ResNet-$50$ backbone with $25$M parameters. Our backbone variant of FFD has only $3$M parameters.}
\label{tab:det_minneapple}

\arrayrulecolor{white!60!black}
\scriptsize
%\footnotesize
%\tiny

\setlength{\tabcolsep}{2.5pt}

\vspace{-0.5ex}

\begin{tabular}{l c c c c c}
\midrule

Method & Backbone (\#Params) & AP &  AP$_{S}$ &  AP$_{M}$  &  AP$_{L}$\\ \midrule

Tile-Faster-RCNN \cite{minneapple} $@$multi-scale & ResNet-$50$ ($25$M)  & $34.1$  & $19.7$ & $51.9$ & $20.8$  \\ 
Faster-RCNN \cite{fasterrcnn} $@$multi-scale & ResNet-$50$ ($25$M)   & $42.3$ & $27.9$ & $58.5$ & $88.2$ \\
YOLO-v$8$ $@$multi-scale \cite{yolov8} & CSPDarkNet ($25$M)   & $44.3$ & $28.3$ & $60.3$ & $84.3$ \\ \midrule
Faster-RCNN \cite{fasterrcnn} $@$single-scale & ResNet-$50$ ($25$M)   & $33.9$ & $15.2$ & $51.2$ & $61.9$ \\ 
YOLO-v$8$ $@$single-scale \cite{yolov8} & CSPDarkNet ($25$M)   & $36.6$  & $18.9$ & $54.7$ & $65.6$ \\ 
DETR \cite{detr}$^\star$ $@$single-scale &  ResNet-$50$ ($25$M)    & $15.2$  & $8.1$ & $19.8$ & $10.6$  \\  \midrule
FFD $@32\times 32$ $@$single-scale & ResNet-$50$ ($25$M)   & $\mathbf{44.6}$  & $\mathbf{29.5}$ & $\mathbf{60.5}$ & $\mathbf{92.2}$  \\
FFD $@32\times 32$ $@$single-scale &  our backbone ($3$M)   & $30.1$  & $21.8$ & $48.8$ & $60.5$  \\ 

 \bottomrule

\end{tabular}
\vspace{-1.0ex}
\end{table}
\subsection{Fastest Training}
Table~\ref{tab:infer} shows the training efficiency on an NVIDIA RTX-$2070$ GPU. Interestingly, FFD has the lowest per-loop training time, primarily attributed to our proposed query assignment and matching strategy, and performing fewer matches relative to the large number of matches in multi-scale detection \cite{fasterrcnn}.
\subsection{Runtime Efficiency Gains}
We report runtime analysis over NVIDIA Jetson Xavier NX, a $10$W palm-sized embedded computing device with $384$ CUDA cores $@$FP$32$ precision. %The runtime can be improved further by at least $2$-$4$ times by using FP$16$, Int$8$ precision and Tensor cores if required. and tensor cores which natively support Int$8$ precision, while the CUDA cores only support FP$32$ and FP$16$ precision. We use only CUDA cores for computations and report the analysis at FP$32$ precision
%
%
%

%%%%%%
%
\begin{table}[t]

\centering
\caption{Training and inference runtime at full precision (FP$32$). FFD-C++ denotes ``C++'' implementation. Training is done on NVIDIA RTX-$2070$, and inference on NVIDIA Jetson Xavier NX.}
\label{tab:infer}

\arrayrulecolor{white!60!black}
\scriptsize
%\footnotesize
%\tiny

\setlength{\tabcolsep}{0.6pt}

\vspace{-0.5ex}

\begin{tabular}{c c c c c c c c}
\midrule

Model & {\tiny Faster-RCNN \cite{fasterrcnn}} & {\tiny SSD \cite{ssd}} & {\tiny FCOS \cite{fcos}} & {\tiny DETR \cite{detr}} & {\tiny YOLO-v$8$ \cite{yolov8}} & {\tiny FFD} & {\tiny FFD-C++} \\ \midrule

Per iteration Training Time & $5.10$s & $4.60$s & $3.40$s & $6.80$s & $0.95$s & $\mathbf{0.70}$s & $\mathbf{0.40}$s \\
Inference $@$FP$32$ & $49$ms & $32$ms & $30$ms & $25$ms & $29$ms & $\mathbf{20}$ms & $\mathbf{11}$ms \\

\bottomrule

\end{tabular}
\vspace{-3.5ex}
\end{table}

\subsubsection{Faster Resource Exemption}
Table~\ref{tab:infer} shows the runtime analysis of different methods with our backbone. It can be seen that FFD is the fastest among all the algorithms. The primary reasons are its minimal architectural components and no multi-scale detection, making FFD a simpler and post-processing-free pipeline.
\par
Runtime is a key metric which determines the duration for which GPU resources shall be held by the detector. From the table, it can be readily seen that FFD has the minimum hold time i.e. $11$ms which is significantly lower than the baselines and is a major achievement and motivation of this work. 
\subsubsection{Resource Allocation to Co-Existing Sub-Systems}
It should be noticed that FFD has a very high speed, but during deployment, images from the sensor/camera can be obtained only at a rate of $30$Hz. However, the high speed ensures the consumption of computing resources for a small duration so that the other compute-intensive algorithms can utilize them. For this reason, even when FFD and other compute-intensive tasks are concurrently running, it does not affect the desired FPS because a lot of computational space is still left on the device. On the other hand, in the existing methods, if two algorithms are deployed simultaneously, each of the algorithms affects the speed of the others because resources are being used for too long. Hence, achieving higher speeds is necessary to guarantee freeing computing resources in a timely manner. Eliminating the post-processing step also reduces the power consumption and programming complexity in contrast to the standard pipelines, which is an additional crucial objective for deployment.
\subsection{Ablation Study}
%

%
%%%%%
%
%%
\begin{table}[t]

\centering
\caption{Effect of synthetic scenes (S.S.) \& augmentation.}
\label{tab:synth}

\arrayrulecolor{white!60!black}
\scriptsize
%\footnotesize
%\tiny

\setlength{\tabcolsep}{7.1pt}

\vspace{-0.5ex}

\begin{tabular}{c c c c c c c c c c c}
\midrule

Colour & Scale & Mirror & Rotate & S.S. & AP &  AP$_{S}$ &  AP$_{M}$ \\ \midrule

 \cmark  &  \cmark &        &        &         & $0.08$  & $0.05$ & $0.2$ \\ 
 \cmark  & \cmark & \cmark &        &         & $8.1$  & $0.9$ & $14.7$ \\ 
 \cmark  & \cmark & \cmark & \cmark &         & $13.6$  & $6.3$ & $15.6$ \\ 
 \cmark  & \cmark & \cmark & \cmark &  \cmark (Ours) & $46.6$  & $31.0$ & $52.1$ \\ \midrule
  \xmark  & \xmark & \xmark & \xmark & \cmark (Ours) & $45.1$ & $34.8$ & $49.2$ \\ 
     \xmark  & \xmark & \xmark & \xmark & \cmark~\cite{semi} & $30.7$  & $31.7$ & $63.9$ \\ 

  \bottomrule

\end{tabular}
\vspace{-3.5ex}
\end{table}
\subsubsection{Synthetic Scenes \& Comprehensive Data Augmentation}
Table~\ref{tab:synth} shows the effect of proposed occlusion-aware scene synthesis along with comprehensive data augmentation on the harvesting region train-test split ($D_h$).
\par
Noticeably, synthetic scenes alone help achieve high accuracy, while using them with data augmentation further improves the performance. Without augmentation, FFD exhibits overfitting, which is intuitive because of the small dataset. Our findings are consistent with \cite{towards}, which mentions the benefits of employing these techniques in the training.
\par
We also compare our scene synthesis technique with the original one \cite{semi} (Table~\ref{tab:synth}). Noticeably, AP decreases for \cite{semi}, which is in accordance with our claim in Sec.~\ref{sec:occ}.
\subsubsection{Tile Size}
The number of queries is determined by the number of predictions per tile ($N_g$) and tile-size. Hence, it is important to see an ablation of how the performance of FFD varies with this parameter. We provide this analysis in Table~\ref{tab:gridsize} by varying the tile-size which is selected such that image resolution can be divided with zero remainder.
\par
We accommodate different tile-sizes by changing the strides in the final stage. For $16 \times 16$ tile-size, the final stage operates at a unit stride, resulting in $H_o=\frac{H}{16}$, $W_o=\frac{W_o}{16}$, while in $64 \times 64$, the last two layers of the final stage operate at a stride $2$, resulting in $H_o=\frac{H}{64}$, $W_o=\frac{W_o}{64}$,
\par
From the experiment, we analyzed that as feature resolution is reduced, accuracy decreases. Accuracy remains stable up to $32 \times 32$ tile-size, and then decreases significantly. While keeping the tile-size to a very small number increases the computations in the backbone for the same number of parameters and more queries. Hence, based on the runtime goals, tile-size can be kept to $32 \times 32$ regardless of the resolution depending upon the requirements.
\subsubsection{Effect of Squeezing Type in \texttt{CCGC}}
\texttt{CCGC} is a crucial component of the \texttt{LOR} module and uses sigmoid by default. However, it is important to analyze the effect of different squeezing activation. We conduct this experiment by replacing sigmoidal activation with softmax operation. 
\par
We observe that softmax faces convergence issues in the same training time (see Table~\ref{tab:squeezingtype}). In addition, from a speed perspective, sigmoid is always faster than softmax since it does not require the normalization step.
%

% %%
\begin{table}[t]

\centering
\caption{Effect of tile-size.}
\label{tab:gridsize}

\arrayrulecolor{white!60!black}
\scriptsize
%\footnotesize
%\tiny

\setlength{\tabcolsep}{6.9pt}

\vspace{-0.5ex}

\begin{tabular}{c c c c c c c}
\toprule

$S(\cdot)$ & \#Params & $N_q$ & Runtime (ms) & AP  &  AP$_{S}$ &  AP$_{M}$\\ \midrule

 $16 \times 16$ & $3.1$M & $1600$ & $14$ms  & $45.3$  & $32.7$ & $50.0$ \\
 $32 \times 32$ & $3.4$M & $800$ & $11$ms  & $46.6$ & $31.0$ & $52.1$  \\
 $64 \times 64$ & $6.5$M  & $400$ &  $13$ms & $10.3$  & $3.1$ & $12.4$\\ 

\bottomrule

\end{tabular}
\vspace{-0ex}
\end{table}
%%%%%%
%
%%

\begin{table}[t]

\centering
\caption{Effect of squeezing type $S(\cdot)$ in \texttt{CCGC}.}
\label{tab:squeezingtype}

\arrayrulecolor{white!60!black}
\scriptsize
%\footnotesize
%\tiny

\setlength{\tabcolsep}{23.0pt}

\vspace{-0.5ex}

\begin{tabular}{c c c c}
\toprule

$S(\cdot)$ & AP &  AP$_{S}$ &  AP$_{M}$\\ \midrule

 Sigmoid  & $46.6$  & $31.0$ & $52.1$ \\
 Softmax & $10.0$ & $0.1$ & $12.5$\\ 

\bottomrule

\end{tabular}
\vspace{-3ex}
\end{table}
%%%%%%
%
%
%%
%
\section{Conclusion}
\label{sec:conc}
This work introduces a Fast-Fruit-Detector (FFD) for UAV-based fruit harvesting tasks in a vertical farming setting. The paper mainly focuses on the visual perceptions system. A deep learning-based single-stage, post-processing free object detector ``FFD'' has been proposed, which can run at $100$FPS on Jetson Xavier NX $@$FP$32$ precision and above $200$FPS $@$FP$16$ or Int$8$. FFD neither requires multi-scale feature fusion to detect small objects nor requires post-processing such as NMS, which is accomplished via novel components of FFD; latent object representation module (\texttt{LOR}), and query assignment and prediction strategy. In addition, we present an approach to generate synthetic scenes to avoid exhaustive manual effort for labelling fruit images. We thoroughly assess FFD on a variety of indoor-outdoor scenes, which suggests that FFD outperforms various mainstream detectors in terms of training-testing efficiency and accuracy evaluation. FFD is not limited only to this purpose, but can be adapted to other robotic applications as well.
\par
\textit{Future direction:} In this work, we have explicitly focused on a single scale. However, given its achievements, FFD holds potential. Hence, it can be extended by fusing multiscale features and also by exploring the use of the recent Transformer-based Mobile backbones, e.g. MobileOne \cite{mobileone}.

%%\nocite{*}
\bibliographystyle{ieeetr}
\bibliography{bibfile}

\end{document}